\documentclass[conference]{IEEEtran}

\usepackage{comment}
\usepackage{cite}
\usepackage{amsmath,amssymb,amsfonts}
\usepackage{algorithmic}
\usepackage{graphicx}
\usepackage{textcomp}
\usepackage{xcolor}
\def\BibTeX{{\rm B\kern-.05em{\sc i\kern-.025em b}\kern-.08em
    T\kern-.1667em\lower.7ex\hbox{E}\kern-.125emX}}
\usepackage{array, colortbl}
\usepackage{ulem, comment}
\usepackage{lscape}
\usepackage{longtable}
\usepackage{adjustbox}
\usepackage{slashbox}
\usepackage{csquotes}
\renewcommand{\mkbegdispquote}[2]{\itshape}
\usepackage{soul,color}
\usepackage{tabularray}
\usepackage[justification=centering]{subfig}
\usepackage[many]{tcolorbox}
\usepackage{balance}
\ifCLASSINFOpdf
\else
\fi
\newcommand{\etal}{\textit{et al.}}

\begin{document}
%
\title{
 Never trust, always verify : a roadmap for Trustworthy AI?}

\author{\IEEEauthorblockN{Lionel Tidjon}
\IEEEauthorblockA{
DGIGL, Polytechnique Montréal\\
Montréal, QC H3C 3A7, Canada\\
}
\and
\IEEEauthorblockN{Foutse Khomh}
\IEEEauthorblockA{
DGIGL, Polytechnique Montréal\\
Montréal, QC H3C 3A7, Canada\\}}


%


\newcommand{\Foutse}[1]{\textcolor{red}{{\it [Foutse: #1]}}}

\maketitle


\begin{abstract}
Artificial Intelligence (AI) is becoming the corner stone of many systems used in our daily lives such as autonomous vehicles, healthcare systems, and unmanned aircraft systems. Machine Learning is a field of AI that enables systems to learn from data and make decisions on new data based on models to achieve a given goal. The stochastic nature of AI models makes verification and validation tasks challenging. Moreover, there are intrinsic biaises in AI models such as reproductibility bias, selection bias (e.g., races, genders, color), and reporting bias (i.e., results that do not reflect the reality). Increasingly, there is also a particular attention to the ethical, legal, and societal impacts of AI. AI systems are difficult to audit and certify because of their black-box nature. They also appear to be vulnerable to threats; AI systems can misbehave when untrusted data are given, making them insecure and unsafe. Governments, national and international organizations have proposed several principles to overcome these challenges but their applications in practice are limited and there are different interpretations in the principles that can bias implementations. In this paper, we examine trust in the context of AI-based systems to understand what it means for an AI system to be trustworthy and identify actions that need to be undertaken to ensure that AI systems are trustworthy. To achieve this goal, we first review existing approaches proposed for ensuring the trustworthiness of AI systems, in order to identify potential conceptual gaps in understanding what trustworthy AI is. Then, we suggest a trust (resp. zero-trust) model for AI and suggest a set of properties that should be satisfied to ensure the trustworthiness of AI systems.
\end{abstract}

\begin{IEEEkeywords}
Artificial Intelligence, Trust, Zero-Trust, Bias, Ethics, Society.   
\end{IEEEkeywords}


%
\IEEEpeerreviewmaketitle

\section{Introduction}

Artificial Intelligence (AI) is achieving a significant success by solving complex challenges in safety-critical Information Technology systems and Operational Technology systems (e.g., programmable logic controllers), that are currently found in healthcare, energy, and transportation infrastructures. AI consists of several fields~\cite{8735821} such as Machine Learning, Deep learning, Natural Language Processing, Computer Vision, Evolutionary and Swarm Intelligence, Knowledge-based Systems, and Multi-Agent Systems. It is showing capabilities close to humans and even higher than humans for some tasks, by reasoning and making decisions from patterns in data. An example of such task is cancer diagnosis. AI can provide early diagnosis and prognosis of various cancer types~\cite{kourou2015machine} such as brain, mammary, and pancreatic cancers. 

With the high growth of AI applications in healthcare, energy, transportation, and finance, there is an increasing attention to the safety, and the ethical, legal, and societal impacts of AI. Microsoft created Tay, a chat bot for people, less than 24 year old and the bot started to tweet racist words within 24 hours after its release~\cite{mitreatlascase}, raising issues about AI fairness. In 2018, a Uber self-driving car killed a pedestrian in Arizona~\cite{uber2018}, raising safety concerns for AI. Robotic surgery was also related to 144 deaths and 1391 injuries from 2000 to 2013~\cite{surgery2015}, raising issues about AI safety and benevolence. Moreover, AI needs more data such as Personal Identifiable Information (PII) and Electronic Health Record (EHR) to improve decisions, causing data privacy and protection issues. AI can be also attacked and fooled by manipulating input data and exploiting software/system flaws~\cite{gao2020backdoor}. Palo Alto Network Security AI research team used domain name mutations to evade a neural network AI-based botnet detector~\cite{mitreatlascase}, raising security concerns for AI. Another well-known ethical issue of AI is its lack of transparency. AI techniques such as Machine Learning and Deep Learning can "accurately" predict outcomes but they fail to provide step-by-step explanation~\cite{hleg_ai} and traceability of how the AI model behaves and how the results are selected. In addition, the accuracy metric is used by most AI experts to evaluate several AI-based systems on test and validation data, after playing with optimization parameters (e.g., learning rate). However, the fact that the AI model works well on test/validation data does not mean it can be trusted on sensitive systems or generalized on real-world systems. In real-time systems, the accuracy of the AI models can increase/decrease according to the quality of the input data and the outcomes become unpredictable, making it uncontrollable; thus causing human damage and financial loss. This leads to the question of AI sustainability. 

The ethical, legal, and societal impacts of AI on the humanity pushed several governments, international and inter-governmental organisations such as European Parliament, Organisation for Economic Co-operation and Development (OECD), and G20 to carefully discuss, analyze, and propose AI ethics principles and guidelines~\cite{tidjon2022different} to prevent AI systems from harming and to make them trustworthy. However, the diversity of the ethical AI principles and their implementation~\cite{tidjon2022different}, gives a feeling that policy makers have a different perception or understanding of the AI trustworthiness. There are also some organisations that follow the concept of "never trust, always verify” known as zero-trust~\cite{laplante2022zero} and prefer to have the full control instead of relying on untrusted AI decision systems. Apart from systems, the trust (resp. zero-trust) is observed in human relationships between people and between organizations. 

\textbf{Contributions.} This paper aims to analyze trust in the context of AI-based systems to understand what it means for an AI system to be trustworthy and identify actions to be undertaken to make AI-based systems trustworthy. This paper provides the following contributions: 
\begin{itemize}
    \item Examination of the different meanings of trust (resp. zero-trust) from the perspective of human to computer (including AI) to get a broad overview of the concepts.
    \item Analysis of the most recent work on the trustworthiness of AI to identify conceptual gaps in understanding the meaning of trustworthiness.
    \item Proposition of a trust (resp. zero-trust) model as well as a set of properties necessary to ensure the trustworthiness of an AI-based system.
\end{itemize}

The rest of this paper is structured as follows. Section~\ref{sota} reviews the relevant literature. Section~\ref{background} describes the different meanings of trust (resp. zero trust) from the perspective of human to computer (including AI). Section~\ref{methodology} describes the methodology of our study. In Section~\ref{evaluation}, we analyze recent trustworthy AI approaches to identify the conceptual gaps in understanding the concept of trustworthy AI. Section~\ref{mitigation} presents a trust (resp. zero-trust) model for AI as well as the ideal properties to ensure the trustworthiness of AI systems. Section~\ref{threats2validity} discusses threats that could affect the validity of the reported results. Section~\ref{conclusion} concludes the paper and outlines some perspectives.

\section{Related work}\label{sota}
This section reviews the related literature on trustworthy AI. 
Davinder \etal{}~\cite{kaur2022trustworthy} defined the needs for a trustworthy AI system, reviewed and compared existing trustworthy AI approaches based on their trustworthy requirements (i.e., Fairness, Explainability, Accountability, Privacy, Acceptance). They also considered human involvement for AI trustworthy (i.e., human before the loop, human in the loop, and human over the loop) and techniques to verify/validate the trustworthiness of AI without compromising its performance such as metamorphic testing, expert panels, benchmarking, and field trials. 

Qinghua \etal{}~\cite{lu2022towards} built a roadmap from the governance-level to system-level to ensure trustworthy AI in software engineering. The roadmap proposes practices during the development and operational process such as requirements engineering (e.g., ethics, traceability), design (e.g., ethical risk, modeling), implementation (e.g., ethical standards, compliance checking), verification and validation (e.g., ethical acceptance testing, formal verification), and operation (e.g., continuous monitoring/validation, accountability).

Schott \etal{}~\cite{thiebes2021trustworthy} defined AI trust concepts as well as principles that should be satisfied by an AI-based system in order to be perceived as trustworthy. The principles include competence/ability, benevolence, integrity, functionality, helpfulness, reliability/predictability, and performance. The principles adopted to make AI trustworthy are beneficence, non-maleficence, autonomy, justice, and explicability.

Wing \etal{}~\cite{wing2021trustworthy} addressed key points to achieve trustworthy AI: trustworthy properties for AI to be extended and formal methods techniques are required. The properties for trustworthy computing are reliability, safety, security, privacy, availability, and usability. The author proposed some additional properties for AI: accuracy, robustness, fairness, accountability, transparency, interpretability/explainability, and ethical. The author also identified some open research challenges that should be addressed to ensure a formal verification of an AI system, in order to increase trust. These challenges include: specification and verification techniques, correctness-by-construction techniques, new threat models and system-level adversarial attacks, processes for auditing AI systems, bias detection, and data de-bias.

Alon \etal{}~\cite{jacovi2021formalizing} defined a trust in the perspective of human and AI via the concept of contracts and formalized the connection between trust and XAI (eXplainable AI). Then, they examined vulnerability in Trust and Trustworthiness, warranted and unwarranted trust, and anticipation of the AI behavior to deviate from trust.

Miles \etal{}~\cite{brundage2020toward} provided a survey of mechanisms (institutional, software, and hardware) to ensure that AI development is conducted in a trustworthy fashion, and proposed recommendations for each mechanism. Institutional mechanisms include third-party auditing, red team exercises, and AI incident sharing. Software mechanisms include audit trails, interpretability, and privacy-preserving machine learning (federated learning, differential privacy, encrypted computation). Hardware mechanisms include hardware security features for AI accelerators and high-precision compute measurement.  

Ben \etal{}~\cite{shneiderman2020human} proposed a human-centered AI framework based on human control and computer control components to increase human performance using properties such as reliability, safety, and trustworthy; and highlight when there are dangers of excessive automation or excessive human control for different applications such as life-critical (trains, airplanes), consumer and professional (recommendation systems, social media platforms). 

Marijn \etal{}~\cite{janssen2020data} provided 12 data governance principles for trustworthy AI including data quality and bias evaluation, changing pattern detection, need to know (i.e., expose only what is necessary), bug bounty, informing people when sharing their data, minimizing authorization for data access, and distributed storage of data. Luciano \etal{}~\cite{floridi2019establishing} also highlighted seven essential principles to achieve trustworthy AI, such as human agency and oversight; robustness and safety; privacy and data governance; transparency; diversity, non-discrimination and fairness; societal and environmental well-being, and accountability. 

LaPlante \etal{}~\cite{laplante2022zero} introduced the concept of zero-trust AI that suggests to trust AI but continuously verify its trustworthiness. They also considered an overhead cost due to the continuous verification of the trustworthiness of AI but outlined the importance of this approach to achieve a trust. In this paper, we extends~\cite{laplante2022zero} with a trust and zero trust models for AI based on ideal properties covering those mentioned in~\cite{wing2021trustworthy, kaur2022trustworthy}.   

\section{Background}\label{background}

In this section, we define what is trust (resp. zero-trust) from human level to AI level. 

\subsection{Human level}

Trust is often perceived as a relation between two entities or more. It has been defined as the willingness of a party (trustor) to be vulnerable to the actions of another party (trustee) based on the expectation that the other will perform a particular action important to the trustor, irrespective of the ability to monitor or control that other part~\cite{mayer1995integrative}. Trust does not mean to be vulnerable or to put a party in a risk position but rather it is a willingness to take risk while engaging with another party. The trust relationship between parties is maintained as long as continuous actions made by both parties are trustworthy (i.e., propensity to trust). However, if actions made by a party increase the vulnerability of another party (i.e., disinclination to trust), the trust relation can be broken depending of some factors (e.g., trust created by contract). Trust contracts outline terms and conditions and clearly identify the parties involved (trustee, trustor). In contractual trust, the trust between trustee and trustor are established by a entity called settlor. The trust contracts can be revocable (i.e., when terms and conditions are broken), or irrevocable (i.e., when the trust relationship is maintained no matter if a party become vulnerable to the other party). 

Continuous trust is important to maintain a trust relationship and it depends on the intrinsic trust of the parties (trustee, trustor), i.e., the internal behaviors, beliefs, and values in which they trust. In~\cite{mayer1995integrative}, the trust beliefs are based on three trust principles: ability, benevolance, and integrity. The ability of a party corresponds to its competencies, group of skills, and characteristics that enable him to take trusted decisions in a specific domain. Benevolence is the belief that a party wants to do good to the other party. Integrity takes place when the trustee adheres to the same principles (personal, moral) in which the trustor believes. 

A party can also opt for zero trust, i.e., to not trust another party. In order to trust the other party, he needs to verify its trustworthiness based on facts, experiences, or policy rules. This verification is done continuously as soon as there is an interaction between parties. This concept, called zero-trust, have been extensively applied in cloud networking and cybersecurity~\cite{rose2020zero}. 

\subsection{Computer level}

In computers, trust applies to hardware and software systems (trustee) as well as their interaction with humans (trustor) and the physical world. Hardware systems have complex operations performed by firmware and multiple chips prone to execution errors and system faults. Users also interact with hardware via software that deals with distributed, concurrency, and usability constraints. Hardware and software systems are also vulnerable to different threats~\cite{8735821} from memory level to CPU level. Thus, trust beliefs in the system context consist of the following additional properties~\cite{wing2021trustworthy}: reliability, safety, security, privacy, availability, and usability. Reliability is concerned about proper execution of the system. Safety ensures that the system does not cause human and cost damages. In order to be trustworthy, the system must be secure and resilient against potential threats. The system often stores private information or its data necessary for operating. This information must be kept private during system execution. For availability, the system must deliver information to users on time at any moment. The system must also be usable by humans. The system trust should increase as experiences of correct execution behaviors increase. In the literature, a computer trust can be also achieved through verification~\cite{wing2021trustworthy}. Formal verification~\cite{berard2013systems} has been used to prove trust properties on specific domains or large domain. Given a model, it checks whether the model satisfies a given property or a set of properties. As a result, a counterexample can be identified, thus provides valuable feedback on how to improve the system. 

At computer level, zero trust considers that no network participant is trustworthy and any access to organisational resources represents a potential threat. Thus, zero trust requires all users, inside/outside of the network of an organization, to be authenticated, authorized, and continuously validated for security configuration and posture before being granted or keeping access to applications and data. It continuously verifies access for all resources and all the time.

\subsection{AI level}

Similar to human trust and computer trust, the trust in AI (trustee) is based on its beliefs or perceptions of its trustworthiness by the user (trustor). The current trustworthiness of the AI system is function of how it is perceived by the user in terms of technical trustworthiness characteristics~\cite{stanton2021trust}. Several organisations~\cite{tidjon2022different} such as G20, European Parliament, General Partnership on AI (GPAI), and Organisation for Economic Co-operation and Development (OECD) have proposed different trustworthiness characteristics (also known as principles). OECD\footnote{https://oecd.ai/en/} promotes five principles to ensure trustworthy AI: inclusive growth, sustainable development and well-being; human-centered values and fairness; transparency and explainability; robustness, security and safety; and accountability. AI must be used for human good and well-being; thus, it should not harm. AI systems must be fair, i.e., they must not discriminate, must be accurate and reliable. In order to be trustworthy, AI systems must be transparent and explainable, i.e., they must have the capability, functionality and features to do what is expected, their executing algorithms can be understood, and they must be perceived by trustors (i.e., users) to be able to achieve user goals. AI systems must also be resilient to threats, that attempt to turn their normal behaviors into harms. In the literature, additional principles have been proposed such as accuracy~\cite{wing2021trustworthy}, acceptance~\cite{kaur2022trustworthy}, predictability and performance~\cite{thiebes2021trustworthy}.

\section{Methodology}\label{methodology}

The \textbf{goal} of this study is to examine trust in the context of AI-based systems to understand what it means for an AI system to be trustworthy, identify actions to be undertaken to make AI-based systems trustworthy, and then propose a trust (resp. zero-trust) model to help bring trustworthy AI in practice. 
The \textbf{perspective} of this study is that of governments, national, and international organisations, which can leverage our findings to guide their AI strategy in developing and implementing trustworthy AI systems. 
The \textbf{context} of this study is a set of 100 documents containing 100 trustworthy properties in the 6 continents (e.g., fairness, transparency), gathered online from reports of national/international organizations. The data sources are selected based on their reliability, recency, and diversity. 
We address the following research questions:
\begin{quote}
	\textbf{RQ$_1$:} \textit{What are the current trustworthy AI properties proposed in the literature ?} This research question aims to review the existing AI trustworthy approaches and identify the current trustworthy properties to guide the planning, design, implementation, and delivery of an AI system.
\end{quote}
\begin{quote}
	\textbf{RQ$_2$:} \textit{What are the gaps in these trustworthy AI properties?} 
	This research question aims to find out where the proposed trustworthy properties may overlap and identify gaps in the defined properties. 
\end{quote}
\begin{quote}
	\textbf{RQ$_3$:} \textit{What are the ideal properties to ensure trustworthy AI ?} This research question aims to propose a model of trust (resp. zero-trust) that describes ideal trustworthy properties to guide the planning, design, implementation, and delivery of an AI system. 
\end{quote}

\begin{table*}[]
\centering
\caption{Trustworthy properties proposed in the literature}
\label{sample-extracted}
\resizebox{\textwidth}{!}{
\scriptsize
\begin{tabular}{|l|l|l|l|l|l|l|l|l|}
\hline
\textbf{Provider}                                                                              & \textbf{Title}                                                                                                                                                                       & Location                                                 & Number & Principles                                                                                                                                                                                                                                                                                                                                                                                                                                  & References                                                                                                                                                              & Date & Type                                                          & Continent     \\ \hline
\begin{tabular}[c]{@{}l@{}}The Institute for \\ Ethical AI \& Machine\\  Learning\end{tabular} & \begin{tabular}[c]{@{}l@{}}The Responsible \\ Machine Learning \\ Principles\end{tabular}                                                                                            & \begin{tabular}[c]{@{}l@{}}United\\ Kingdom\end{tabular} & 8      & \begin{tabular}[c]{@{}l@{}}Human augmentation, \\ Bias evaluation, \\ Explainability By \\ Justification, Reproducible \\ operations, \\ Displacement strategy, \\ Pratical Accuracy, Trust \\ by privacy, Data risk \\ awareness\end{tabular}                                                                                                                                                                                              & \begin{tabular}[c]{@{}l@{}}https://ethical.institute\\ /principles.htmtl\end{tabular}                                                                                   &      & \begin{tabular}[c]{@{}l@{}}Research\\  Institute\end{tabular} & Europe        \\ \hline
\begin{tabular}[c]{@{}l@{}}Universite de \\ Montreal\end{tabular}                              & \begin{tabular}[c]{@{}l@{}}Montréal Declaration\\ for Responsible AI\end{tabular}                                                                                                    & Canada                                                   & 10     & \begin{tabular}[c]{@{}l@{}}well-being, autonomy, \\ intimacy and privacy, \\ solidarity, democracy, \\ equity, inclusion, caution,\\  responsibility and \\ environmental sustainability\end{tabular}                                                                                                                                                                                                                                       & {\color[HTML]{000000} \begin{tabular}[c]{@{}l@{}}https://www.montreal\\ declaration-responsible\\ ai.com/reports-of-montreal\\ -declaration\end{tabular}}               & 2017 & University                                                    & North America \\ \hline
IEEE                                                                                           & \begin{tabular}[c]{@{}l@{}}Ethically Aligned Design:\\  A Vision for Prioritizing \\ Human Well-being with \\ Autonomous and Intelligent\\ Systems - General Principles\end{tabular} & US                                                       & 5      & \begin{tabular}[c]{@{}l@{}}Human Rights;Prioritizing \\ Well-being;Accountability;\\ Transparency;A/IS Technology \\ Misuse and Awareness of It\end{tabular}                                                                                                                                                                                                                                                                                & {\color[HTML]{000000} \begin{tabular}[c]{@{}l@{}}https://standards.ieee.org/\\ content/dam/ieee-standards/\\ standards/web/documents/\\ other/ead\_v2.pdf\end{tabular}} & 2018 & \begin{tabular}[c]{@{}l@{}}Inter\\ national\end{tabular}      & North America \\ \hline
India Government                                                                               & Ethics and Human Rights                                                                                                                                                              & India                                                    & 5      & \begin{tabular}[c]{@{}l@{}}equality, safety \& reliability, \\ inclusivity \& non-discrimination, \\ transparency, accountability and \\ privacy \& security\end{tabular}                                                                                                                                                                                                                                                                   & {\color[HTML]{000000} \begin{tabular}[c]{@{}l@{}}https://indiaai.gov.in/\\ research-reports/responsible\\ -ai-part-1-principles-for-\\ responsible-ai\end{tabular}}     & 2021 & Government                                                    & Asia          \\ \hline
AI Forum                                                                                       & Trustworthy AI in Aotearoa                                                                                                                                                           & \begin{tabular}[c]{@{}l@{}}New \\ Zealand\end{tabular}   & 5      & \begin{tabular}[c]{@{}l@{}}Fairness and Justice;Reliability, \\ Security and Privacy;Transparency\\ ;Human Oversight and \\ Accountability;Well being\end{tabular}                                                                                                                                                                                                                                                                          & {\color[HTML]{000000} \begin{tabular}[c]{@{}l@{}}https://data.govt.nz/assets/\\ data-ethics/algorithm/\\ Trustworthy-AI-in-\\ Aotearoa-March-2020.pdf\end{tabular}}     & 2020 & Government                                                    & Ocenia        \\ \hline
Research ICT Africa                                                                            & \begin{tabular}[c]{@{}l@{}}Recomendations on the \\ inclusion subSaharan Africa\\  in Global AI Ethics\end{tabular}                                                                  & \begin{tabular}[c]{@{}l@{}}South \\ Africa\end{tabular}  & 6      & \begin{tabular}[c]{@{}l@{}}Introduce safeguards to balance AI\\ opportunities and risks; Protect individual\\  and collective privacy rights in crossborder\\  data flows; Define African values for AI \\ and align AI frameworks with such values; \\ Practise fair and socially-responsible AI; \\ Build inclusive partnerships based on \\ community and cocreation; Adopt an \\ adaptive, open minded\\ and humble approach\end{tabular} & {\color[HTML]{000000} \begin{tabular}[c]{@{}l@{}}https://researchictafrica.net/\\ wp/wp-content/uploads/2020/\\ 11/RANITP2019-2-AI-Ethics.pdf\end{tabular}}             & 2019 & \begin{tabular}[c]{@{}l@{}}Research \\ Institute\end{tabular} & Africa        \\ \hline
\begin{tabular}[c]{@{}l@{}}Chile's Ministry \\ of Science\end{tabular}                         & \begin{tabular}[c]{@{}l@{}}Cross-cutting AI \\ principles\end{tabular}                                                                                                               & Chile                                                    & 4      & \begin{tabular}[c]{@{}l@{}}IA with a focus on people's well-being, \\ respect for human rights and security; \\ IA for sustainable development; \\ Inclusive AI; Globalized \\and evolving AI\end{tabular}                                                                                                                                                                                                                                    & {\color[HTML]{000000} \begin{tabular}[c]{@{}l@{}}https://www.carey.cl/en/\\ chile-presents-its-first-\\ national-artificial-intelligence\\ -policy/\end{tabular}}       & 2021 & Government                                                    & South America \\ \hline
\begin{tabular}[c]{@{}l@{}}The high-level expert group on \\ artificial intelligence (AI HLEG)\\  and European AI Alliance\end{tabular}                                     & \begin{tabular}[c]{@{}l@{}}The Assessment List For Trustworthy\\ Artificial Intelligence (ALTAI) \\ For Self Assessment\end{tabular}                                                 &                      & 7                       & \begin{tabular}[c]{@{}l@{}}human agency and oversight; technical robustness \\ and safety; privacy and   data governance; transparency;\\ diversity, non-discrimination and fairness; \\ societal and environmental well-being; \\ and accountability\end{tabular}                                                                                                                                                                                                                    & {\color[HTML]{000000} \begin{tabular}[c]{@{}l@{}}https://ec.europa.eu/newsroom/dae/\\ document.cfm?doc\_id=68342\end{tabular}}                                                                                                                           & 2019         & International & Europe             \\ \hline
\begin{tabular}[c]{@{}l@{}}Organisation de coopération et de  \\ développement économiques (OECD)\end{tabular}                                                              & OECD AI Principles                                                                                                                                                                   &                      & 5                       & \begin{tabular}[c]{@{}l@{}}Inclusive growth, sustainable development and \\ well-being;Human-centred values and fairness;\\ Transparency and explainability;Robustness, \\ security and   safety;Accountability\end{tabular}                                                                                                                                                                                                                                                          & {\color[HTML]{000000} {\ul https://oecd.ai/en/ai-principles}}                                                                                                                                                                                            & 2019          & International & 6 continents       \\ \hline
IEEE                                                                                                                                                                        & \begin{tabular}[c]{@{}l@{}}Ethically Aligned Design: A \\ Vision for Prioritizing Human \\ Well-being with Autonomous and\\ Intelligent Systems -\\  General Principles\end{tabular} & US                   & 5                       & \begin{tabular}[c]{@{}l@{}}Human Rights;Prioritizing Well-being;\\ Accountability;Transparency;A/IS Technology \\ Misuse and Awareness of It\end{tabular}                                                                                                                                                                                                                                                                                                                             & {\color[HTML]{000000} \begin{tabular}[c]{@{}l@{}}https://standards.ieee.org/content/dam/\\ ieee-standards/standards/web/documents\\ /other/ead\_v2.pdf\end{tabular}}                                                                                     & 2018          & International & North America      \\ \hline
\begin{tabular}[c]{@{}l@{}}Colombia’s National Planning \\ Department,  together with the ICT\\ Ministry (MinTIC) and the President’s \\ administrative Office\end{tabular} & \begin{tabular}[c]{@{}l@{}}Ethical Framework For Artifical \\ Intelligence In Colombia\end{tabular}                                                                                  & Colombia             & 10                      & \begin{tabular}[c]{@{}l@{}}Transparency;Explanation;Privacy;Human control of the \\ decisions of an artifcial intelligence system \\ (Human-in-the-loop and Human-over-the-loop);\\ Security;Responsibility;Non-discrimination;Inclusion;\\ Prevalence of the rights of children and adolescents;\\ Social Benefit;\end{tabular}                                                                                                                                                      & {\color[HTML]{000000} \begin{tabular}[c]{@{}l@{}}https://dapre.presidencia.gov.co/dapre\\ /SiteAssets/documentos/ETHICAL\%20\\ FRAMEWORK\%20FOR\%20ARTIFICIAL\\ \%20INTELLIGENCE\%20IN\\ \%20COLOMBIA.pdf\end{tabular}}                                  & 2020          & Governement   & South America      \\ \hline
Microsoft                                                                                                                                                                   & Microsoft AI principles                                                                                                                                                              & US                   & 6                       & \begin{tabular}[c]{@{}l@{}}Fairness; Reliability \& Safety;Privacy \& Security;\\ Inclusiveness; Transparency;Accountability\end{tabular}                                                                                                                                                                                                                                                                                                                                             & {\color[HTML]{000000} \begin{tabular}[c]{@{}l@{}}https://www.microsoft.com/en-us\\ /ai/responsible-ai\end{tabular}}                                                                                                                                      &             & Company       & North America      \\ \hline
IBM                                                                                                                                                                         & Foundational properties for AI ethics                                                                                                                                                & US                   & 5                       & Explainability;Fairness;Robustness;Transparency;Privacy                                                                                                                                                                                                                                                                                                                                                                                                                               & {\color[HTML]{000000} {\ul https://www.ibm.com/cloud/learn/ai-ethics}}                                                                                                                                                                                   &              & Company       & North America      \\ \hline
Google                                                                                                                                                                      & Google AI principles                                                                                                                                                                 & US                   & 7                       & \begin{tabular}[c]{@{}l@{}}Be socially beneficial;Avoid creating or reinforcing \\ unfair bias;Be built and tested for safety;Be \\ accountable to people;Incorporate privacy design \\ principles;Uphold high standards of scientific excellence; \\ Be made available   for uses that accord with \\ these principles\end{tabular}                                                                                                                                                  & {\color[HTML]{000000} {\ul https://ai.google/principles/}}                                                                                                                                                                                               &               & Company       & North America      \\ \hline
\begin{tabular}[c]{@{}l@{}}Australia's Department of Industry,   \\ Science, Energy and Resources\end{tabular}                                                              & \begin{tabular}[c]{@{}l@{}}Australia’s Artificial Intelligence \\ Ethics Framework / Australia's \\ AI Ethics principles\end{tabular}                                                & Australia            & 8                       & \begin{tabular}[c]{@{}l@{}}Human, societal and environmental wellbeing;\\ Human-centred values;Fairness;Privacy protection \\ and security;Reliability and safety;Transparency and \\ explainability;Contestability;Accountability\end{tabular}                                                                                                                                                                                                                                       & {\color[HTML]{000000} \begin{tabular}[c]{@{}l@{}}https://www.industry.gov.au/data\\ -and-publications/australias-artificial\\ -intelligence-ethics-framework\\ /australias-ai-ethics-principles\end{tabular}}                                            & 2018         & Government    & Oceania            \\ \hline
Indian Government                                                                                                                                                           & Ethics and Human Rights                                                                                                                                                              & India       & 5                       & \begin{tabular}[c]{@{}l@{}}equality, safety \& reliability, inclusivity \& \\ non-discrimination, transparency, accountability \\ and privacy \& security\end{tabular}                                                                                                                                                                                                                                                                                                                & {\color[HTML]{000000} \begin{tabular}[c]{@{}l@{}}https://indiaai.gov.in/research-reports/\\ responsible-ai-part-1-principles\\ -for-responsible-ai\end{tabular}}                                                                                         & 2021         & Government    & Asia               \\ \hline
\begin{tabular}[c]{@{}l@{}}Israel's Ad hoc Committee on Artificial\\ Intelligence (CAHAI)\end{tabular}                                                                      & \begin{tabular}[c]{@{}l@{}}Harnessing Innovation: Israeli \\ Perspectives on AI Ethics and \\ Governance : Six Ethical Principles \\ for AI\end{tabular}                             & Israel              & 6                       & \begin{tabular}[c]{@{}l@{}}Fairness (equality, biases prevention, discrimination \\ prevention, avoidance of education and socioeconomic\\  gaps); Accountability (Transparency,   Explainability, \\ Ethical and legal responsibility);Protecting human right\\ (Bodily integrity, Privacy, Autonomy, Civil and political \\ rights);Cyber and   information security;Safety \\ (internal, external);Maintaining a competitive market\end{tabular}                                   & {\color[HTML]{000000} \begin{tabular}[c]{@{}l@{}}https://sectech.tau.ac.il/sites/sectech.tau.ac.il\\ /files/CAHAI\%20-\%20Israeli\\ \%20Chapter.pdf\end{tabular}}                                                                                        & 2020          & Government    & Asia               \\ \hline
Government of Canada                                                                                                                                                        & \begin{tabular}[c]{@{}l@{}}Responsible use of artificial\\  intelligence (AI)\end{tabular}                                                                                           & Canada               & 5                       & \begin{tabular}[c]{@{}l@{}}understand and measure;be transparent;provide \\ meaningful explanations; be as open as we can;\\ provide sufficient training\end{tabular}                                                                                                                                                                                                                                                                                                                 & {\color[HTML]{000000} \begin{tabular}[c]{@{}l@{}}https://www.canada.ca/en/government\\ /system/digital-government/digital\\ -government-innovations\\ /responsible-use-ai.html\#toc1\end{tabular}}                                                       & 2017          & Government    & North America      \\ \hline
\begin{tabular}[c]{@{}l@{}}Asilomar Conference on Beneficial AI \\ (Future of Life Institute)\end{tabular}                                                                  & ASILOMAR  AI Principles                                                                                                                                                              & US                   & 13                      & \begin{tabular}[c]{@{}l@{}}Safety;Failure Transparency;Judicial Transparency;\\ Responsibility;Value Alignment;Human Values;\\ Personal Privacy;Liberty and Privacy;Shared Benefit;\\ Shared Prosperity;Human Control;Non-subversion;\\ AI Arms Race\end{tabular}                                                                                                                                                                                                                     & {\color[HTML]{000000} \begin{tabular}[c]{@{}l@{}}https://futureoflife.org/2017/08\\ /11/ai-principles/\end{tabular}}                                                                                                                                     & 2017          & International & North America      \\ \hline
\begin{tabular}[c]{@{}l@{}}United Arab Emirates - Smart \\ Dubai   initiative\end{tabular}                                                                                  & Dubai AI principles                                                                                                                                                                  & United Arab Emirates & 4                       & \begin{tabular}[c]{@{}l@{}}Ethics (fair, accountable, explainable, transparent), \\ Security (safe, secure, controllable), Humanity (AGI+\\ superintelligence, human values, benfeficial use of AI), \\ Inclusiveness (human values, freedom, dignity, respect \\ privacy, benefit of AI, governance of AI)\end{tabular}                                                                                                                                                              & {\color[HTML]{000000} \begin{tabular}[c]{@{}l@{}}https://www.digitaldubai.ae\\ /initiatives/ai-principles\end{tabular}}                                                                                                                                  & 2019          & Government    & Asia               \\ \hline
UNI Global Union                                                                                                                                                            & \begin{tabular}[c]{@{}l@{}}Top 10 Principles For Ethical \\ Artificial Intelligence\end{tabular}                                                                                     & Switzerland         & 10                      & \begin{tabular}[c]{@{}l@{}}Demand That AI Systems Are Transparent;Equip AI \\ Systems With an “Ethical   Black Box”;Make AI \\ Serve People and Planet;Adopt a Human-In-Command \\ Approach;Ensure a Genderless, Unbiased AI;Share \\ the Benefits of AI   Systems;Secure a Just Transition \\ and Ensuring Support for Fundamental Freedoms \\ and Rights;Establish Global Governance Mechanisms;\\ Ban the Attribution of Responsibility to Robots;\\ Ban AI Arms Race\end{tabular} & {\color[HTML]{000000} \begin{tabular}[c]{@{}l@{}}http://www.thefutureworldofwork.org/media\\ /35420/uni\_ethical\_ai.pdf\end{tabular}}                                                                                                                   & 2017         & International & 6 continents       \\ \hline
\begin{tabular}[c]{@{}l@{}}European Commission \\ (DG Research and Innovation)\end{tabular}                                                                                 & \begin{tabular}[c]{@{}l@{}}Ethics By Design and Ethics of Use\\ Approaches for Artificial Intelligence\end{tabular}                                                                  &                      & 6                       & \begin{tabular}[c]{@{}l@{}}Respect for Human Agency (autonomy, dignity, freedom); \\ Privacy and Data governane;Fairness;Individual, Social and \\ Environmental Well-being;Transparency;Accountability and \\ Oversight\end{tabular}                                                                                                                                                                                                                                                 & {\color[HTML]{000000} \begin{tabular}[c]{@{}l@{}}https://ec.europa.eu/info/funding-tenders\\ /opportunities/docs/2021-2027/horizon/\\ guidance/ethics-by-design-and-ethics-of-use\\ -approaches-for-artificial-\\ intelligence\_he\_en.pdf\end{tabular}} & 2021          & International & Europe             \\ \hline
Accenture                                                                                                                                                                   & Responsible AI principles                                                                                                                                                            &Irelande            & 5                       & \begin{tabular}[c]{@{}l@{}}Accountability;Transparent;Addressing Bias;\\ Humand and Machine; Data and   Security\end{tabular}                                                                                                                                                                                                                                                                                                                                                         & {\color[HTML]{000000} \begin{tabular}[c]{@{}l@{}}https://www.switch.ch/export/sites/default\\ /procure/.content/.files/Why-RAI-Matters\\ -Pinar-Wennerberg.pdf\end{tabular}}                                                                             & 2021          & International & Europe             \\ \hline
Government of New South Wales                                                                                                                                               & \begin{tabular}[c]{@{}l@{}}Artificial Intelligence (AI) Ethics \\ Policy: Mandatory Ethical Principles\\  for the use of AI\end{tabular}                                             & Australia            & 4                       & Fairness;Privacy and security;Transparency;Accountability                                                                                                                                                                                                                                                                                                                                                                                                                             & {\color[HTML]{000000} \begin{tabular}[c]{@{}l@{}}https://www.digital.nsw.gov.au/policy/\\ artificial-intelligence-ai/artificial-intelligence\\ -ai-ethics-policy/mandatory-ethical\end{tabular}}                                                         & 2020          & Government    & Oceania            \\ \hline
PwC                                                                                                                                                                         & \begin{tabular}[c]{@{}l@{}}The General Ethical AI Principles \\ + The epistemic principles\end{tabular}                                                                              & UK                   & 9                       & \begin{tabular}[c]{@{}l@{}}Accountability;Data privacy;Lawfulness and compliance; \\ Beneficial AI;Humann agency;Safety;Fairness; \\ Interpretability (Explainability, Transparency, Provability)\\ ; Reliability, Robustness, Security\end{tabular}                                                                                                                                                                                                                                  & {\color[HTML]{000000} \begin{tabular}[c]{@{}l@{}}https://www.pwc.com/gx/en/issues\\ /data-and-analytics/artificial-intelligence\\ /what-is-responsible-ai.html\end{tabular}}                                                                             &               & Company       & Europe             \\ \hline
\end{tabular}
}
\end{table*}

\subsection{Data collection}

Three selection criteria have been defined for data collection: reliability, 
more recent data, and diversity. 
The reliability criteria ensures that the data can be trusted or is provided by a trusted sources. More than 100 reports and websites from companies, national, and international organizations have been inspected to collect trustworthy AI properties. For each candidate report, we assessed the trustworthiness of the institutions behind the report. The recency criteria ensures that the data source contains recent information about trustworthy AI properties. The 100 reports have been published between 
2017 and 2022. To select a report, we check if it contains a 
clear reference to trustworthy AI properties.

The diversity criteria is used to ensure that we collect data 
from different countries and continents. The final set of reports analyzed in this paper
covers countries from Europe, North America, South America, Oceania, Africa, and Asia. The references of these reports can be found in our replication package \cite{tech-report-v1}.

\subsection{Data extraction}

In the hundred reports, we have manually selected 100 trustworthy AI principles randomly, representing a confidence level of 90\% with an error margin of 8.25\%. Each principle contains 2 to 15 keywords including  Fair, Fairness, Secure, Security, Responsible, Responsibility, Inclusive, Inclusiveness, Safe, Safety, Transparent, Transparency, Robust, Robustness, Explainable, and Explainability.

Based on Microsoft Excel, we have built regular expressions to extract these string keywords. The following regular expression was used to extract each trustworthy keyword: \textit{*Safe*} (Safety), \textit{*Transparen*} (Transparency), \textit{*Robust*} (Robustness), \textit{*Explainab} (Explainability), \textit{*Fair*} (Fairness), \textit{*Secur*} (Security), \textit{*Responsib*} (Responsibility), and \textit{*Inclusive*} (Inclusiveness). For each property, we extracted and recorded the following information: 
the name of the report's provider (e.g., Canada Government), the title of the report, the location where the report is published, the number of trustworthy keywords contained in the report, the contained trustworthy properties, the reference to get the properties, the date when it was released, the type of institution (e.g., government, international, company, university), and continent. An example of recorded information 
is shown in Table~\ref{sample-extracted}. The complete dataset is available in our replication package \cite{tech-report-v1}. 

We have extracted trustworthy keywords from the following countries: Canada, United States (US), United Kingdom (UK), China, Australia, Singapore, United Arab Emirates (UAE), Spain, Switzerland, Chile, Colombia, Sweden, Belgium, Israel, Amsterdam, New Zeland, India, Germany, Hong Kong, South Korea, South Africa, Finland, Japan, France, Netherlands, Norway, Russia, Ireland, and Italy. The trustworthy AI keywords include: Fairness, Transparency, Accountability, Autonomy, Human Dignity, Responsibility, Safety, Security, Robustness, Explainability, Interpretability, Well-being, Human Oversight, Human Rights, Sustainability, Equity, Reliability, Privacy, Justice, Solidarity, Beneficence, Non-maleficence, Non-discrimination, Contestability, Human-centric, Inclusiveness, Trustworthy, Democracy, Governance, Bias, Integrity, Controllability, and Accuracy.

After extraction, we count the number of occurrences of trustworthy properties to analyze their prevalence in the literature and identify gaps in the properties.

\section{Analysis of trustworthy AI approaches}\label{evaluation}

In order to answer to RQ1, we define two sub research questions: What are the current trustworthy AI approaches ? and What are the trustworthy AI properties used ? 

\subsection{What are the current trustworthy AI approaches ?}

Schott \etal{}~\cite{thiebes2021trustworthy} adopted the following principles: beneficence, non-maleficence, autonomy, justice, and explicability. In~\cite{wing2021trustworthy}, J. Wing~\etal{} also proposed several trustworthy AI properties including accuracy, robustness, fairness, accountability, transparency, interpretability/explainability, and ethical. 

Floridi~\etal{}\cite{floridi2018ai4people} proposed five ethical AI principles (i.e., beneficence, non-maleficence, autonomy, justice, explicability) to lead the development and adoption of AI for the benefit of society and presented 20 concrete recommendations to assess, develop, incentivise, and support good AI. However, the proposed recommendations are high-level and mainly focus on AI ethics governance. This work does not evaluate the application of AI ethics in practice. 

Jobin~\etal{}\cite{jobin2019global} performed an in-depth analysis of AI guidelines and observed that they are structured around five global principles (i.e., transparency, justice and fairness, non-maleficence, responsibility, and privacy). They also identified two important actions to be undertaken by the global community: (1) translating principles into practice and (2) seeking harmonization between AI ethics codes (soft law) and legislation (hard law). 

Fjeld~\etal{}\cite{fjeld2020principled} proposed a visualization of AI principles classified by themes (e.g., International Human Rights, Privacy) and sectors (multi-stakeholder, inter-governmental, private, government, civil society) to provide a high-level snapshot of the current state of AI ethics governance. Nelson \etal{}~\cite{nelson2020toolkit} recommends to normalize, organize, and operationalize racial equity throughout data integration. They suggest an ongoing process at each stage of the data life cycle-planning, i.e., data collection (e.g., shared data, inclusive data), data access (open data, restricted data, available data), use of algorithms and statistical tools (i.e., following Responsibility, Explainability, Accuracy, and Auditability, Fairness principles), data analysis (e.g., using participatory research to bring multiple perspectives to the interpretation of the data), reporting and dissemination (e.g., providing clear documentation of the data analysis process with analytic files for reproducibility).

The AI High Level Expert Group (HLEG)~\cite{smuha2019ethics} proposed technical and non-technical methods to implement AI ethics. Technical methods include architectures for Trustworthy AI, ethics and rules of law by design (e.g., transparency-by-design, fairness-by-design), explanation methods (e.g., IBM XAI), testing and validation (e.g., red teaming, bug bounty programs). Non technical-methods include regulation, codes of conduct, standardization, certification, accountability via governance frameworks, diversity and inclusive design teams, education and awareness to foster an ethical mind-set, stakeholder participation and social dialogue.

\subsection{What are the trustworthy AI properties used in the literature ?} \label{trust-out}

\begin{figure}[h]
\centering
\includegraphics[scale=0.61]{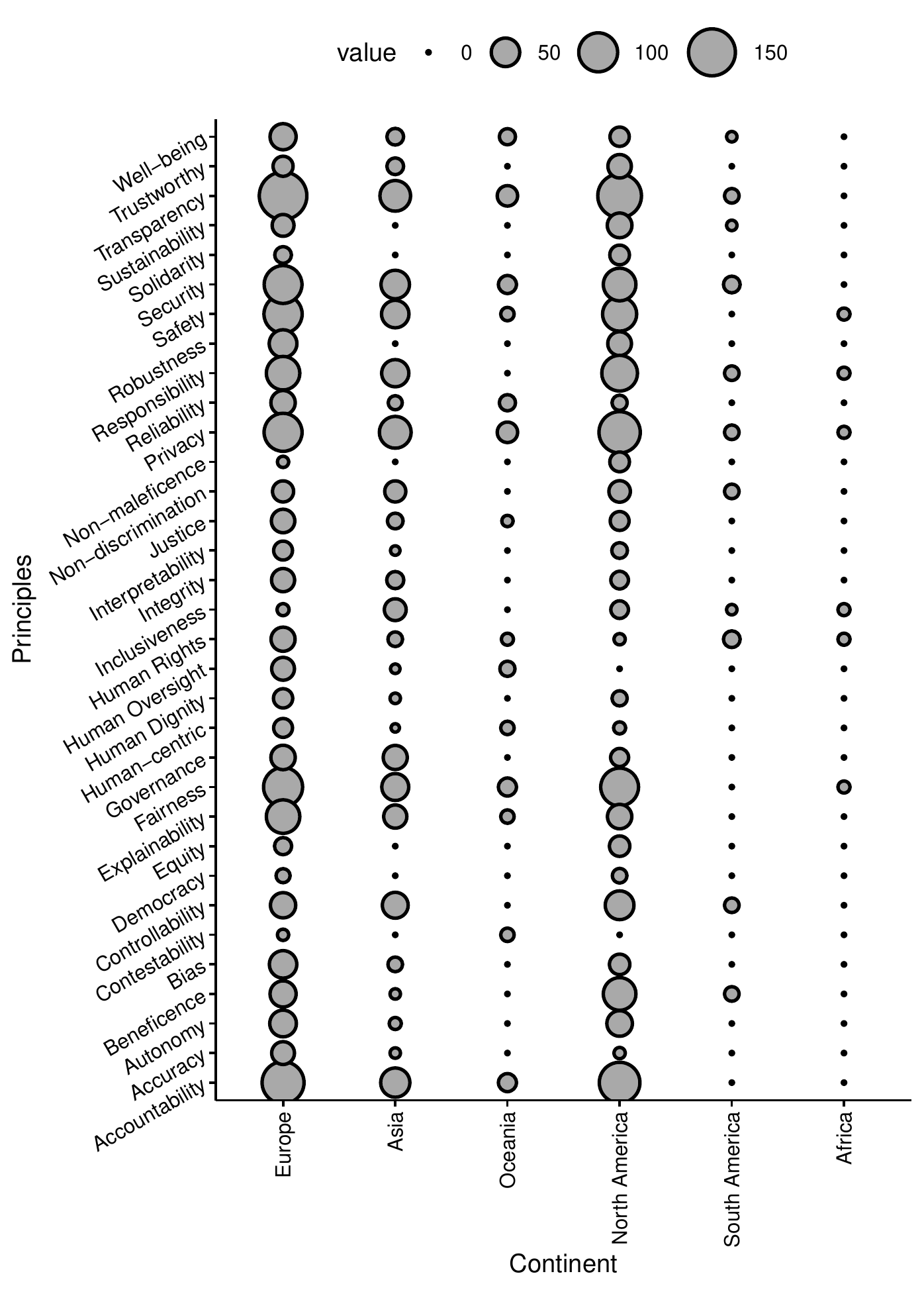}
\caption{Number of occurrences per principle per continent}
\label{fig:nbrprinccont1}
\end{figure}

This research question aims to identify the trustworthy properties used in the literature. Fig.~\ref{fig:nbrprinccont1} shows that Human Rights and Privacy properties appear in all continents. Transparency, Fairness, Safety and Security principles have the highest number of occurrences and they appear in 5 continents: Asia, Europe, North America, Oceania, South America, and Africa. Responsibility, Well-being, and Inclusiveness principles are also found in 5 continents: North America, Asia, Europe, South America, Oceania, and Africa. Accountability and Explainability principles appear in 4 continents with a high number of occurrences: Europe, North America, Asia, and Oceania. 

\begin{figure*}[h]
\centering
\includegraphics[scale=0.54]{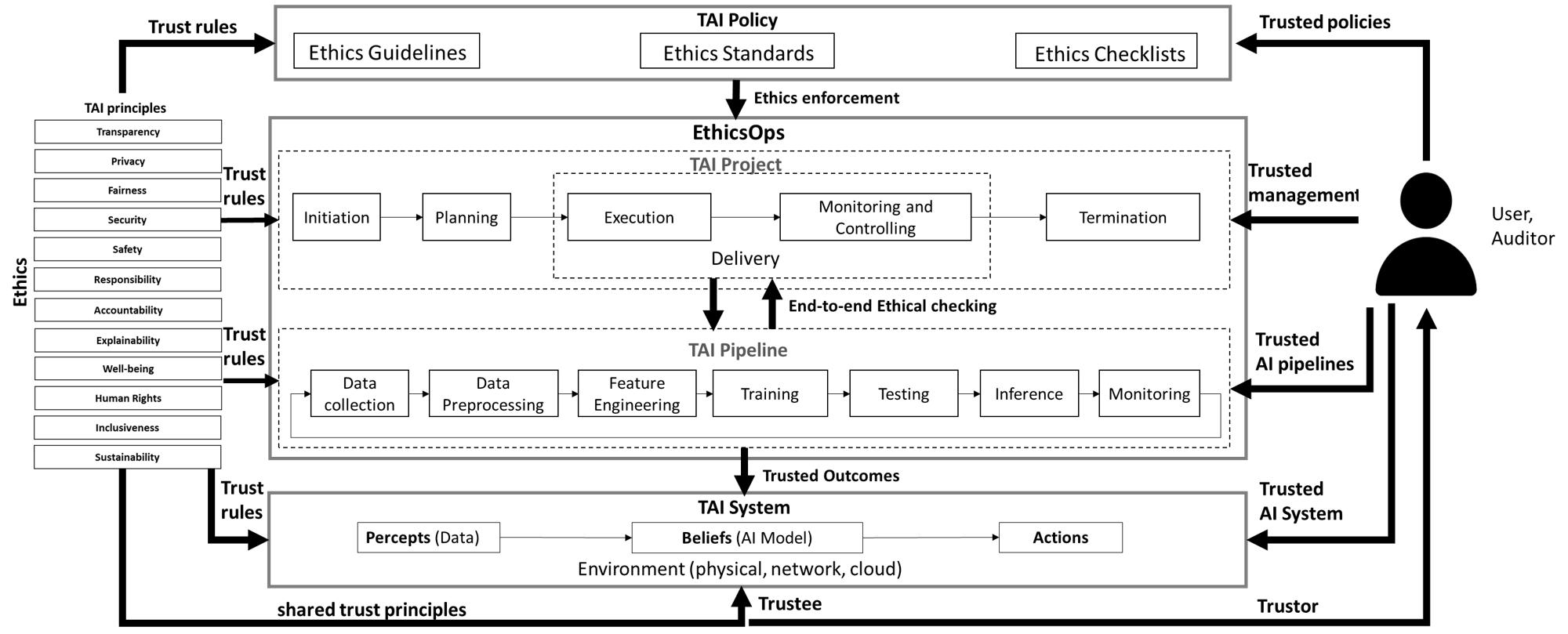}
\caption{The proposed Trust AI model}
\label{trust-model}
\end{figure*}

\subsection{What are the gaps in these trustworthy AI properties ?}\label{sustain-c}

In order to answer RQ2, we analyze overlaps and gaps in the trustworthy properties. After analysis of the different trustworthy AI properties, we observe that different organizations and governments have a different perception of trustworthy AI; since some principles are cited in some continents and not in others (see Section \ref{trust-out}). In Fig.~\ref{fig:nbrprinccont1}, we observe overlapping properties such as Human-centric and Human dignity and Human Rights, Equity and Justice, Beneficence and Well-being and Non-maleficence. Europe and North America put an emphasis on Non-maleficence, while it is not the case in Asia, Oceania, South America, and Africa. However, Asia, Oceania, and South America highlighted Well-being property. Non-maleficence is included in Well-being. Human centered property include Human dignity and Human Rights. Justice property include Equity. Beneficence, Well-being and Non-maleficence properties are similar. Beneficence also means Non-maleficence. Beneficence and Non-maleficence are included in Well-being.

After analysis, we also observe that Explainability and Accountability properties are missing in South America and Africa. Responsibility and Inclusiveness properties are missing in Oceania. Safety property is missing in South America. Sustainability property are missing in Oceania, Asia, and Africa.

\section{Trustworthy AI Model}\label{mitigation}

This section aims to propose ideal properties to achieve trustworthy AI as stated in RQ3. In~\cite{tidjon2022different}, we have identified 11 global trustworthy AI principles that cover most continents with a high number of occurrences: Transparency, Privacy, Fairness, Security, Safety, Responsibility, Accountability, Explainability, Well-being, Human Rights, and Inclusiveness. In addition to the 11 principles, we add the Sustainability principle based on gaps identified in Section~\ref{sustain-c}. The sustainable AI property focuses on sustainable data sources, power supplies, and infrastructures to measure and reduce the carbon footprint from training and/or tuning an AI algorithm. These 12 principles are used as ideal properties to define our trust (resp. zero-trust) model and they can be enforced during the end-to-end lifecycle of an AI project from the planning phase to the delivery phase (i.e., deployment); and  throughout the maintenance and evolution  
of an AI system. 
(see Fig.~\ref{trust-model}).

\begin{figure*}[h]
\centering
\includegraphics[scale=0.53]{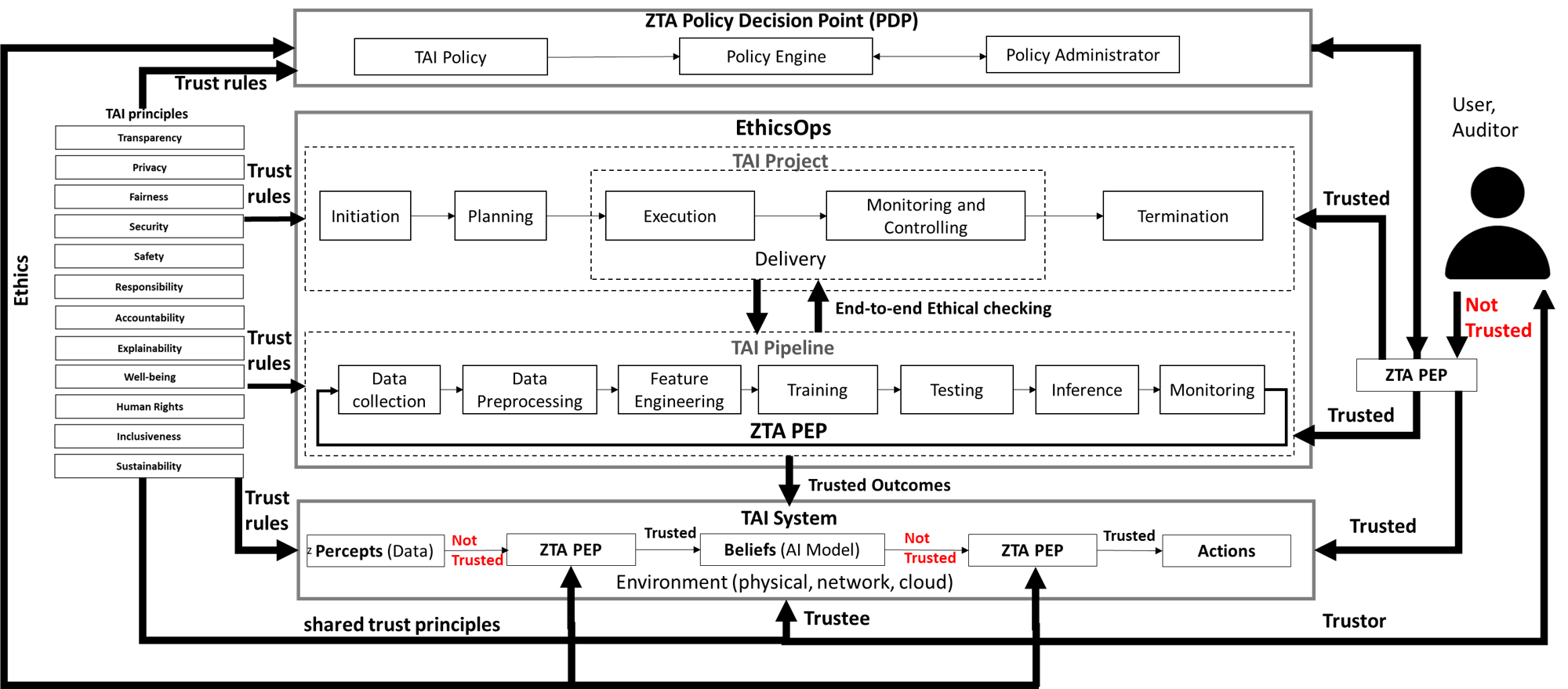}
\caption{The proposed Zero-Trust AI model}
\label{zero-trust-model}
\end{figure*}

\subsection{Trust AI Model}

The Trust AI model consists of five components: human (trustor), AI system (trustee), Trustworthy AI (TAI) principles, TAI policy, and EthicsOps. The trustor can be a user of the system or an auditor. He trusts the AI system when actions performed by it are trustworthy and they share the same TAI principles. These TAI principles are ruled in the TAI policies of the organization. TAI policies consist of ethics guidelines, standards, and checklists~\cite{tidjon2022different} used to guide, assess and control the application of ethical TAI principles from the governance level to the technical level. These policies must be enforced in the AI project lifecycle using EthicsOps. EthicsOps is the continuous planning, execution and control of ethical TAI principles during the AI project lifecycle and the related AI pipelines. It ensures that business requirements are pro-ethically aligned (i.e., follows TAI principles), ethical controls are enforced (i.e., application of TAI policies), and that ethics by design practices~\cite{eu-ethos} are applied during the AI project lifecycle. During initiation and planning phases, business requirement specifications (e.g., functional, non-functional, architecture) must follow TAI policies. Team selection must take into accounts inclusiveness property irrespective of the race, gender, skills, and disability. During the execution and delivery phases, ethics by design practices~\cite{eu-ethos} can be used to help prevent ethical issues using Key Ethics Indicators (KEI) metrics such as transparency~\cite{mitchell2019model,ribeiro2016should,chen2018traceability,mora2021traceability},  privacy~\cite{abadi2016deep, sweeney2002k, mohassel2017secureml}, fairness~\cite{corbett2018measure,10.1145/3457607},  security~\cite{bonawitz2017practical,papernot2016towards,mohassel2017secureml}, explainability~\cite{ribeiro2016should,lundberg2017unified}, and sustainability~\cite{schwartz2020green}. 

AI pipelines must also follow EthicsOps to ensure end-to-end trust from data collection to monitoring. Training data must not contain bias~\cite{10.1145/3457607} and must be private and secure. Data protection~\cite{scarfone2008sp, csa_report} can be enforced \textit{in transit} by using TLS encryption, \textit{at rest} by using AES-based keys, and \textit{in use} by dynamic analysis of the training data. Right identity and access management (IAM) based on least privilege~\cite{mccarthy1800identity} can be used to restrict access to training data, models and its parameters, and prediction results. The source code of the AI pipelines can be regularly scanned to identify vulnerabilities to maintain AI pipelines in a trust state. The trained models must also be resilient to adversarial threats~\cite{tramer2020adaptive} to ensure a trustworthy AI system. The outcomes or artifacts produced from the AI pipelines (e.g., models, binaries) must be trustworthy.

In production, the whole AI system (trustee) in place must be trusted by the trustor (user, auditor). Usually, a system consists of percepts, beliefs, effects/actions, and the environment. For an AI system, percepts are consumed data (e.g., images) from the environment by the system in order to be processed internally. An environment can be physical (e.g., phone, house) and virtual (e.g., computer, network, cloud). These perceived data as well as the underlining environment must be trustworthy. Beliefs are probabilistic predictions made by AI models operating inside the AI system. Actions are performed by the AI system based on its beliefs. EthicsOps ensures the continuous monitoring of these actions to identify any deviation in trust so that failing or untrusted components can be updated to move the system into trust state. Verification and testing~\cite{berard2013systems} can also be used to enforce AI trust using model checking techniques on AI model parameters~\cite{urban2021review}.   

\subsection{Zero-Trust AI Model}

The proposed Zero-trust AI (ZTA) model extends the trust model with a verification layer applied at each phase of the AI project and its related AI pipelines, and each trust interaction between trustee and trustor. Zero-trust AI model consists of six components: human (trustor), AI system (trustee), Trustworthy AI (TAI) principles,
Policy Enforcement Point (PEP), Policy Decision Point (PDP), and EthicsOps (see fig.~\ref{zero-trust-model}). 

The Zero-Trust AI (ZTA) PEP enables, monitors, and closes trust relationship between the trustor and the trustee. ZTA PEP interacts with a policy administrator~\cite{rose2020zero} responsible for establishing and/or stop trust relationship between parties. To take the appropriate decision, the policy administrator can reason based on a decision engine (i.e., the policy engine). The policy engine manages the decision to authorize access to a resource (e.g., data, predictions) for a given subject (e.g., models, users). To do so, the policy engine uses TAI policy based on ethical trustworthy AI principles, guidelines, standards, and checklists in order to verify the trustworthiness of the entity. ZTA PDP takes an immediate decision based on instructions received from the policy administrator in the ZTA PEP component. The verification of the trustworthiness of the AI models can be achieved using VeriDeep~\cite{he2018verideep}, DeepZ~\cite{singh2018fast}, and other tools such as RefineZono and RefinePoly~\cite{urban2021review}.

\section{Threats to validity}\label{threats2validity}

In this work, we have analyzed 100 reports, and extracted 100 set of trustworthy AI principles. This process was done manually and we used basic Excel functions to extract trustworthy keywords. Thus, we may have missed information containing relevant keywords. It took 3 weeks to search, read, 
identify, extract, and record relevant information from the 100 reports. 

In addition, there were some trustworthy principles with long size that may contain redundant keywords and it can affect the number of occurrences of a trustworthy keyword for a given continent. However, we have used distinct keywords to avoid redundancy and we have also verified document sources and the online publishers to ensure that they are trustworthy. During the extraction process, we have repeated statistics 2 times to avoid any miscalculation. 

\section{Conclusion}\label{conclusion}

In this work, we have examined existing trustworthy AI approaches and identified the trustworthy properties highlighted in the literature. After analysis of the properties, we identified gaps in the different properties as well as some overlaps. Based on these gaps, we propose a trust (resp. zero trust) model to ensure a trustworthy AI. Results show that Transparency is the most cited principle among the studied countries. The 12 ideal trustworthy properties identified  are Transparency, Privacy, Fairness, Security, Safety, Responsibility, Accountability, Explainability, Well-being, Human Rights, Inclusiveness, and Sustainability. The proposed trust and zero-trust models are based on EthicsOps to ensure end-to-end application and control of trustworthy AI principles during the AI project lifecycle. In the future, we plan to develop some ethical tools and propose practical assessment mechanisms based on the trust and zero-trust models.
\section*{Acknowledgment}
This work is partly funded by the Fonds de Recherche du Québec (FRQ), Natural Sciences and Engineering Research Council of Canada (NSERC), Canadian Institute for Advanced Research (CIFAR), and Mathematics of Information Technology and Complex Systems (MITACS).



%
\bibliographystyle{IEEEtran}
\bibliography{main}

\end{document}